# A Robust Framework for Deep Learning Approaches to Facial Emotion Recognition and Evaluation

Nyle Siddiqui
Department of Computer Science
University of Wisconsin – Eau Claire
Eau Claire, US
siddiqun8701@uwec.edu

Thomas Reither
Department of Computer Science
University of Wisconsin – Eau Claire
Eau Claire, US
reithert2424@uwec.edu

Rushit Dave
Department of Computer Science
University of Wisconsin – Eau Claire
Eau Claire, US
daver@uwec.edu

Dylan Black
Department of Computer Science
University of Wisconsin – Eau Claire
Eau Claire, US
blackdt0976@uwec.edu

Tyler Bauer
Department of Computer Science
University of Wisconsin – Eau Claire
Eau Claire, US
bauerta7262@uwec.edu

Mitchell Hanson
Department of Computer Science
University of Wisconsin – Eau Claire
Eau Claire, US
hansonmk0604@uwec.edu

*Abstract*—Facial emotion recognition (FER) is a vast and complex problem space within the domain of computer vision and thus requires a universally accepted baseline method with which to evaluate proposed models. While test datasets have served this purpose in the academic sphere, real-world application and testing of such models lacks any real comparison. Therefore, we propose a framework in which models developed for FER can be compared and contrasted against one another in a constant, standardized fashion. A lightweight convolutional neural network (CNN) is trained on the AffectNet dataset – a large, variable dataset for facial emotion recognition - and a web application is developed and deployed with our proposed framework as a proof of concept. The CNN is embedded into our application and is capable of instant, real-time facial emotion recognition. When tested on the AffectNet test set, this model achieves an accuracy of 55.09% for emotion classification of eight different emotions. Using our framework, the validity of this model and others can be properly tested by evaluating a model's efficacy not only based on its accuracy on a sample test dataset, but also on in-the-wild experiments. Additionally, our application is built with the ability to save and store any image captured or uploaded to it for emotion recognition, allowing for the curation of more quality and diverse facial emotion recognition datasets.

## I. INTRODUCTION

The superior performance of deep learning algorithms has been observed across the many domains within computer science. Computer vision [1], game AI [2], and natural language processing [3] serve as a few examples of areas that have been revolutionized by the introduction and widespread use of machine learning. Specifically, user authentication and general cyber security protocols have witnessed a dramatic increase in the utilization of machine and deep learning for improved performance [4, 5]. Physical and behavioral biometrics have been proposed as the two leading modalities in both accuracy and reliability for user authentication. Despite this growth in popularity, there is no established baseline, real-world environment in which the performance and subsequent results of models trained for user authentication using physical biometrics can be tested. While previously published results contain the general evaluation metrics of a model after being rigorously trained and tested on a dataset, these results may not be truly indicative of a particular model's performance when applied to a real-world scenario.

Thus, this paper proposes a robust framework to test the performance of various facial emotion recognition models when implemented in a real-world environment. Our framework was built as a facial emotion recognition web application, in which users can take a live photo from their webcam or upload an image of their face and view which emotions were detected to be present. These emotions were detected using a CNN trained on the AffectNet dataset [6], however, any pre-trained model for facial emotion recognition can use our framework to evaluate its efficacy when presented with out-of-distribution data. While quality, balanced datasets exist for facial emotion recognition – such as AffectNet - the training and testing sets created from these datasets are not sufficiently large nor diverse enough to be compared to the variability that arises when real users interact with these models with live, never-before-seen images. Therefore, the novel contributions of this paper are as follows

- Propose a framework that allows for the robust evaluation of models trained for facial emotion recognition and the future compilation of a new facial emotion dataset by saving all uploaded user images
- Train and evaluate a CNN using the AffectNet dataset and compare the results between alternatively proposed models in the literature
- Develop and foster a standardized method to evaluate the efficacy of deep learning models when applied to real-world scenarios

II. RELATED WORK

The synergetic nature between easily available big data and high performing deep learning algorithms has strongly influenced a plethora of novel solutions to notoriously difficult problems. User authentication techniques [7-13], data security in autonomous vehicles [14], and the general domain of IoT [15-17] are but a few examples of areas in computer science that have capitalized on the previously unattainable benefits of deep learning. Algorithms that fall under the blanket term deep learning, specifically the model family of neural networks, possess a superior ability to extract additionally granular features from raw data compared to more traditional statistical methods; for example, recurrent neural networks in the field of natural language processing, convolutional neural networks in image processing and object detection, and the use of neural networks in deep reinforcement learning and game theory. However, the performance of these algorithms are usually heavily dependent on the quality and quantity of data that is available to train on and thus pose a significant obstacle for problem spaces that lack sufficient amounts of data or are relatively new.

Currently, there are a variety of publicly available datasets for facial emotion recognition, such as EMOTIC [18], the Extended Cohn-Kanade Dataset (CK+) [19], FER-2013 [20, 21], and the aforementioned AffectNet, among many more. As with many similar problems in the domain of computer vision, convolutional neural networks have recently excelled when assigned the task of facial emotion recognition. This is mostly due in part to the increased availability of large amounts of facial emotion data with which to train these deep learning models. Obtaining a standardized, large amount of raw data is usually the first obstacle to overcome when applying deep learning to a relatively new field, such as facial emotion recognition.

The CK+ and FER-2013 datasets, introduced in [19] and [20], respectively, both established strong baseline evaluation benchmarks in the relative infancy of facial emotion recognition and deep learning. The CK+ dataset contains thousands of images of centered, forward-facing, well-lit human faces that portray one of seven emotions: neutral, happy, surprise, sad, fear, disgust, and anger. The FER-2013 dataset was created with images collected from a large image search on the Google search engine. This resulted in 35,000 images that contain human faces with various poses, lighting, features, and occlusion. The diversity and inconsistency present in this dataset may seem like a hindrance when training CNNs for FER, however it better equips these models to detect human emotion at a level equal to or superior to humans [22-24].

[21] exhibits the state-of-the-art performance that CNNs are capable of achieving when properly trained and tuned. The authors note that the heterogeneity and variability of both human faces and the surrounding scenery they occupy create an immense difficulty in accurate and reliable facial emotion recognition. Thus, the CNN's superior ability in raw feature extraction and image processing made it a prime candidate to test on the FER-2013 dataset given the variable nature of the dataset. Using a VGGNet architecture with meticulously fine-tuned hyperparameters, the authors of [21] were able to achieve an accuracy of 73.28% on the FER-2013 dataset, outperforming the estimated human performance accuracy of 65.5% [20]. Furthermore, the transparency of CNNs, or lack thereof colloquially known as the "black box" nature of CNNs, is often called into question as a significant and often dangerous downside of the use of deep learning. Understanding and visualizing how information is encoded inside the CNN as raw data propagates through the network is necessary in order to effectively evaluate and improve CNN performance. Thus, [21] used saliency maps - which utilize the properties of backpropagation to highlight which pixels contributed the most to certain emotion predictions - to allow for more granular and targeted changes when the CNN was performing irregularly as well as increase general data visualization. Thus, our framework was equipped with the ability to display each model's confidence in each emotion classification to improve model interpretability.

III. METHODOLOGIES

A. AffectNet

The AffectNet dataset has superseded alternative datasets as one of the best due to its high quantity and quality of facial emotion images. The full, complete dataset contains more than one million images collected from the internet using three major querying search engines. Unfortunately, the full dataset is not publicly available even for academic research due to the difficulty of manually annotating and verifying each image and its corresponding ground truth emotion. Consequently, the roughly 300,000 images available have been manually annotated and verified to represent eight discrete facial emotions: neutral, happy, sad, surprise, fear, disgust, anger, and contempt. The dataset consists of images each with 224x224 pixels and three RGB channels. Similarly to FER-2013, this dataset was also curated using a search engine and consequently, the nature of the dataset is highly heterogenous as each image varies in pose, lighting, angle, occlusion, color, and much more. Furthermore, the valence and arousal intensity of each image is also provided to allow for facial emotion recognition through different modalities. For example, this paper solely uses a CNN to process and map raw pixel data directly to a facial emotion, however methods that include using facial landmarks (similar to priors in object detection) have also been proven as viable. When introducing this dataset, the authors also provided results from two baseline deep neural networks they had trained to classify human emotions and found that they excelled at both categorical and dimensional emotion recognition. These two models are currently being used to automate the labelling process; however these images are still reviewed by humans for verification and are not included in the dataset disseminated for academic use. The provided training and testing set consisted of roughly 287,651 and 4,000 images, respectively.

### B. CNN architecture and data preprocessing

Despite the high performance and reliability of state-of-the-art CNNs, their immense consumption of time and resources is not conducive to developing a general-use system that operates in real-time. Thus, following [24], we implemented a lightweight AlexNet variant CNN to achieve a balance between accuracy and runtime. AlexNet's [25] novel contribution was primarily the time-efficient application of CNNs to high-resolution image recognition and was first observed in the famous ImageNet Large-Scale Visual Recognition Challenge (ILSVRC). AlexNet's use of decrementing convolutional kernel sizes, starting from 9x9 and reducing to 3x3, in addition to the shallower and narrower model size proposed in [24] aided in reducing in computational overhead. The Adam optimizer was used with its recommended parameters of $\alpha = 0.001$, $\beta_1 = 0.9$, $\beta_2 = 0.999$, and $\epsilon = 10^{-8}$, which are thought to be the optimal default values for general deep learning training on large datasets.

To reduce model weights and training time, each image has its RGB values normalized to values between 0 and 1. Furthermore, to mitigate frequency bias, we utilize class weights for each emotion present in the dataset. Because there are 134,915 instances of "happy" yet 4,250 instances of "contempt" in the dataset, class weighting is imperative to reduce the model's bias towards predicting more commonly occurring emotions and also increase general accuracy. The preprocessing method used in this paper differs from [24], as we omit cropping all images to size 128x128 using the facial bounding boxes provided by AffectNet. Because this model was being trained with the intent of implementation in a pseudo real-world environment in our system, it was integral that the model was invariant towards perturbations surrounding a face in an image. However, it should be noted that images taken or uploaded by the user may vary depending on a variety of factors, such as the user's webcam dimensions, the original size of an uploaded picture, or even the centering of the user's face before taking a picture. Thus, we include a customizable bounding box that the user may interact with after taking or uploading a picture to allow the user more flexibility; for example, centering their own face or highlighting only one face in a picture with many faces. Regardless of the size of the box the user chooses, the cropped image will be resized to 224x224 prior to being evaluated by the model. Fig. 1 exhibits sample real world classifications by the model of the two most frequent and least frequent emotions in the AffectNet dataset. More details regarding the user interface are discussed in the next section.

There are five convolutional "blocks" inside the CNN, consisting of a convolutional layer, a 2x2 max-pooling layer, and a batch normalization layer, as seen in Table I. The convolutional and max-pooling layers are integral to the model's ability to process images, while the batch normalization layer serves to normalize value after each convolutional pass, thereby reducing model weights and training time. The final convolutional block leads to the emotion recognition section of the model, which starts with a flatten layer that reduces the output of the final convolutional block into a single-dimensional vector. Since this model only processes images of a static 224x224 size, the resulting outputted vector size is always 2048. The flatten layer is followed by two fully connected dense layers, which lead to a final dense layer with a softmax activation – the only layer in the model that does not have a ReLU activation function. This layer contains 8 neurons for each emotion and outputs the model's emotion prediction. Our model is trained on the AffectNet dataset and reached a peak test accuracy rate of 0.5509 after training for 13 epochs. Human annotation performance for emotion classification on the AffectNet dataset is thought to be around 60%, exhibiting the notable performance of our model.

### C. Framework Architecture

FERS, or the facial emotion recognition system, was built as a web application to achieve maximum accessibility for our users. The user interface, as seen in Fig. 2, was built using React - a component-based JavaScript framework. The user has the ability to either take a live picture of themselves using a connected webcam, or to upload a JPEG or PNG file. After an image has been evaluated by the model the top-3 emotions detected with their corresponding confidence value are shown to the user. While top-1 accuracy is a stricter and more rigorous evaluation metric, top-3 accuracy does provide more transparency and detail to the user with regards to the model's emotion classification prediction. We also implemented a back-end and database using Netlify and Google's Firebase services, respectively, to collect and store any images uploaded or taken by the user. As the use of the web application increases, the number of images saved will soon become sufficiently large to create a novel facial emotion detection dataset. As mentioned in Section II, the datasets that often lead to effective FER contain images that vary in many image properties (pose, lighting, etc.) Thus, the independent and variable image collection process within our system is a great candidate for the curation of future datasets.

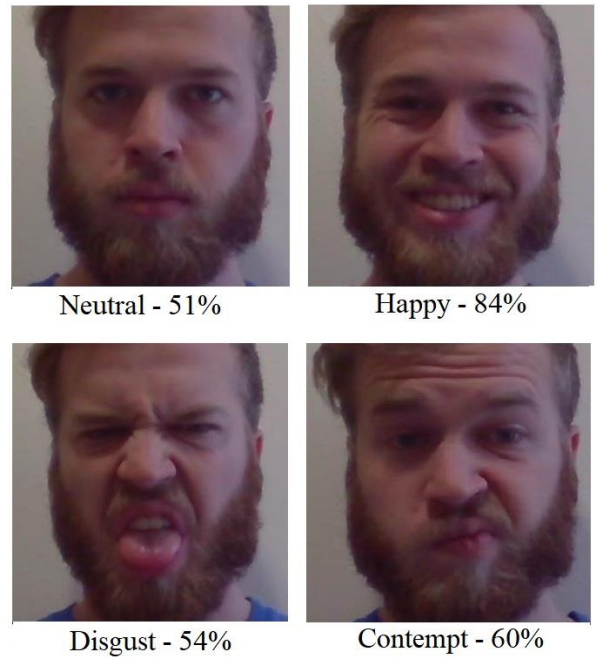

Figure 1. Real world examples of emotions correctly classified by the model with confidence values

TABLE I. MODEL ARCHITECTURE

| Layer | Shape |
|---|---|
| Convolutional | 9x9x16 |
| Max-Pooling | 2x2 |
| Convolutional | 7x7x32 |
| Max-Pooling | 2x2 |
| Convolutional | 5x5x64 |
| Max-Pooling | 2x2 |
| Convolutional | 3x3x128 |
| Max-Pooling | 2x2 |
| Convolutional | 3x3x128 |
| Max-Pooling | 2x2 |
| Flatten | 2048 |
| Dense (ReLU) | 1024 |
| Dense (ReLU) | 1024 |
| Dense (Softmax) | 8 |

The facial emotion recognition model is embedded directly in the web application using the TensorFlow.js library, allowing for instant emotion recognition. After the model has been fully trained, we use the TensorFlow.js library to convert a saved TensorFlow model into the correct file format to embed into our application. Herein lies the flexibility of our system – any fully trained facial emotion detection model (built in TensorFlow) can be saved, converted, and embedded into our application for easy evaluation and testing. Currently, outside of provided test sets in FER datasets, methods to evaluate the performance of facial emotion recognition models in a real-world environment are sparse. Therefore, our novel system produces a rigorous environment in which to test both past and future deep learning-related facial recognition models.

IV. DISCUSSION AND ANALYSIS

Our approach to facial emotion recognition is that of a categorical approach; by using a CNN for raw image processing and feature extraction, we operate under the assumption that each emotion is separate and distinct from one another and will therefore be distinguishable to a properly trained CNN. However, dimensional approaches to FER - in which differences between emotions are thought to be continuous in nature - rely more on strong facial landmarks. These methods have been shown to be effective for FER [27] and exploit valence and arousal to ameliorate the dimensional representation of an emotion in an image. Table II compares the results of alternative models trained on the AffectNet dataset to the results observed in this article, with green indicating that our model had a higher accuracy and red indicating that the alternative model had a higher accuracy on the AffectNet test dataset. Machine learning has expectedly shown promising results in with regards to image processing and FER, and as such a support vector regression (SVR) algorithm was tested alongside other CNN and RNN-based models in [26]. The AlexNet variant model we used in this paper had a higher accuracy rate on the AffectNet test set than nearly all of the models, with the few remaining higher performing models shown in Table II. Despite our model's superior performance on the test data, there is still no baseline method available to evaluate its functionality as a real-world application; hence our scrutiny on not only the implementation of this model but also proposing a framework in which to evaluate it as well. Further, other deep learning models proposed in [24] observed slightly higher test accuracies on the AffectNet dataset. These models are compatible with our framework and can thus be embedded in our web application and immediately tested in our standardized, constant environment.

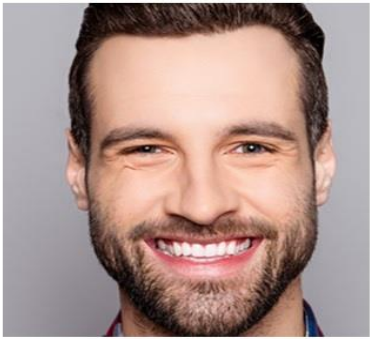

Figure 2. An example of facial emotion recognition using our framework with a stock image. The top-3 detected emotions are listed

Class weighting did seem to mitigate frequency bias as a dramatic test accuracy increase of 8-10% was observed when introduced, however the vast difference in number of instances for frequent emotions (i.e. happy, neutral, angry) versus infrequent emotions (i.e. contempt, disgust) may be impossible to fully overcome with our categorical approach. Even with AffectNet's colossal number of available high quality training images, the imbalance of emotion frequency highlights the flaw of using a singular dataset for model evaluation. Therefore, the adoption of our framework can lead to both broader model evaluation methods as well as the curation of a more balanced yet diverse dataset with regards to AffectNet.

## V. LIMITATIONS

The developmental stage of our framework was a success as observed in our functional web application, however utilizing our framework to test alternative FER models proposed in the literature has not been exercised yet. Furthermore, our application is capable of storing every image uploaded or captured in order to compile an extensive FER dataset similar to AffectNet, yet due to its relative infancy, this functionality has not been fully realized. Further analysis of other FER models through the use of our framework will naturally solve this problem and can lead to the development of another quality FER dataset.

TABLE II. ACCURACY COMPARISONS ACROSS DIFFERENT MODELS.

| Model | Peak Accuracy on AffectNet |
|---|---|
| Our Model | 0.5509 |
| VGGNet Variant [24] | 0.58 |
| MobileNet Variant [24] | 0.58 |
| SVR [6] | 0.277 |
| CNN [6] | 0.470 |
| 2Att-CNN [26] | 0.487 |
| 2Att-Mt [26] | 0.539 |
| 2Att-2Mt [26] | 0.635 |

Additionally, we use a smaller model size in this paper to aid in the real-time operation of our web application, but at the cost of accuracy. If real-time operation is not imperative to a certain FER system, deeper CNN models can be similarly trained and fine-tuned on the AffectNet dataset to possibly improve performance and reduce generalization error due to the model's increased capacity. [28] explores the use of these deeper models and verifies their ability to perform at an even higher level than the model used in this paper or proposed in [24]. Further research in the optimization of deep neural networks in general could also contribute to more efficient runtimes, thereby increasing accuracy. Lastly, confidence values for detected emotions are calculated using the softmax outputs of each neuron in the final layer. While still relatively efficient, improved methods of confidence calculation without the use of potentially misleading softmax values have been proposed in other domains of computer vision [29, 30].

## VI. CONCLUSION

In this article, a versatile and robust framework has been proposed for the rigorous testing and evaluation of FER models in a pseudo real-world environment. A web application was designed using our framework as a proof of concept. Our application allows for simple and intuitive interactions with the proposed framework and expedites the potential curation of a novel FER dataset. Moreover, a computationally lightweight CNN trained on the AffectNet dataset was implemented in our web application to test real-time operation and as a baseline comparison for future model testing. This lightweight model achieved a peak test accuracy of 0.5509 on the AffectNet dataset, operating at near human-level efficacy. Other models, such as VGGNet variant architectures, have achieved higher accuracy rates and thus indicate the need for further research in the optimal implementation of deep learning for facial emotion recognition.

Intelligent behavior analysis could revolutionize aspects of security in a multitude of business and governmental areas, however the intrusive nature of this research must also be closely monitored and regulated. While it may not be necessary in the early stages of development, note that all users on our web application must create a profile and consent to the privacy policy informing them of the process of data collection and storage.

ACKNOWLEDGMENT

We would like to thank the Blugold Supercomputing Cluster at the University of Wisconsin – Eau Claire for providing the computational hardware to train our model.